\documentclass{article}

\usepackage{arxiv}
\usepackage{natbib}
\usepackage[utf8]{inputenc} % allow utf-8 input
\usepackage[T1]{fontenc}    % use 8-bit T1 fonts
\usepackage{hyperref}       % hyperlinks
\usepackage{url}            % simple URL typesetting
\usepackage{booktabs}       % professional-quality tables
\usepackage{amsmath}        % align, \text, \notag, etc.
\usepackage{amssymb}        % \varnothing and other symbols
\usepackage{amsthm}         % theorem-like environments (definition)
\usepackage{mathtools}      % \coloneqq
\usepackage{amsfonts}       % blackboard math symbols
\usepackage{nicefrac}       % compact symbols for 1/2, etc.
\usepackage{microtype}      % microtypography
\usepackage{lipsum}
\usepackage{graphicx}
\graphicspath{ {./figures/} }

\title{Adaptive Graph-of-Islands Evolution for Automatic Feature Engineering with LLMs}

\author{
 Sha Li \\
  Department of Computer Science\\
  Virginia Tech\\
  Blacksburg, VA 24060 \\
  \texttt{shal@vt.edu} \\
   \And
 Naren Ramakrishnan\\
Department of Computer Science\\
  Virginia Tech\\
  Alexandria, VA 22305\\
  \texttt{naren@cs.vt.edu} \\
}
\usepackage{latexsym}
\usepackage{xspace}
\usepackage[ruled,vlined]{algorithm2e}
\SetKwFor{ForPar}{for}{do in parallel}{end}

\def\topo{\textsc{TopoFE}\@\xspace}

\usepackage{multirow}      % \multirow
\DeclareMathOperator*{\argmin}{arg\,min}
\DeclareMathOperator*{\argmax}{arg\,max}
\newtheorem{definition}{Definition}

\newcommand{\cF}{\mathcal{F}}
\newcommand{\cL}{\mathcal{L}}
\newcommand{\cM}{\mathcal{M}}
\newcommand{\cP}{\mathcal{P}}
\newcommand{\cG}{\mathcal{G}}

\newcommand{\bX}{\mathbf{X}}

\newcommand{\by}{\mathbf{y}}
\newcommand{\R}{\mathbb{R}}
\newcommand{\E}{\mathbb{E}}
\newcommand{\Perf}{\mathrm{Perf}}

\begin{document}
\maketitle
\begin{abstract}
Automatic feature engineering (AutoFE) for tabular data requires discovering informative transformations from a combinatorially large and heterogeneous program space. Existing approaches suffer from three fundamental limitations: classical methods rely on fixed operator libraries with limited expressivity, LLM-based methods generate proposals from static prompts without retaining prior search experience, and evolutionary methods employ fixed migration policies that ignore task-specific cross-family transfer utility. We introduce \topo, a framework that formulates AutoFE as graph-structured multi-island evolutionary program search. The transformation space is partitioned into semantically coherent families, each explored by a dedicated island through LLM-guided mutation and crossover. Each island maintains a Prompt Adaptation Memory that accumulates accept/reject feedback to steer future proposals toward productive regions of the search space without parameter updates. To coordinate global exploration, \topo dynamically learns a directed topology graph whose edge weights encode empirical transfer utility between transformation families. Cross-island transfer is triggered by adaptive saturation detection and performed through LLM-mediated hybrid synthesis, enabling discovery of compositional feature programs that cannot emerge from isolated local search. Experiments on 29 tabular datasets show that \topo consistently outperforms most state-of-the-art AutoFE methods on both classification and regression tasks. Beyond predictive performance, \topo produces feature sets with lower redundancy and higher representational coverage, while the learned topology graph acquires meaningful task-specific transfer structure correlated with downstream gains. The discovered feature programs also transfer reliably across diverse downstream predictors and LLM backbones, demonstrating that the improvements arise from \topo’s structured search and adaptive coordination mechanisms instead of backbone-specific generation capability.
\iffalse

Automatic feature engineering (AutoFE) for tabular learning can be naturally formulated as a program synthesis problem, where the objective is to discover predictive feature transformations from an exponentially large search space. Recent advances in large language models (LLMs) have expanded the expressiveness of AutoFE by enabling feature program generation beyond predefined operator libraries. However, existing LLM-based approaches remain fundamentally limited by stateless generation and homogeneous search: feature proposals are produced from static prompts without accumulating search experience, while single-population exploration quickly converges to dominant transformation patterns and rarely discovers complementary feature compositions across transformation families. We propose \topo, a topology-aware multi-island evolutionary framework for LLM-guided feature engineering. \topo combines family-specialized exploration, adaptive prompt memory, and topology-guided knowledge transfer to efficiently discover diverse and compositional feature programs. Experiments on 29 public tabular datasets demonstrate consistent improvements over state-of-the-art AutoFE methods across classification and regression tasks. Beyond predictive performance, \topo discovers more diverse and transferable feature programs that generalize across multiple downstream predictors and LLM backbones.
\fi
\end{abstract}

% keywords can be removed
%\keywords{First keyword \and Second keyword \and More}

\section{Introduction}
\label{sec:intro}
Feature engineering (FE) is the process of transforming raw variables into predictive representations that expose the latent structure of a learning problem \citep{dong2018feature}, it remains one of the most consequential yet least principled stages of the tabular machine learning pipeline. In tabular settings, where data are heterogeneous, semantically rich, and often laden with domain-specific structure, the quality of engineered features frequently determines model performance more than architectural choices or hyperparameter tuning~\citep{zhang2023openfe, hollmann2023large}. Constructing high-value features requires more than syntactic enumeration. For example, a clinically meaningful risk ratio, a financial volatility indicator, or a behavioral frequency feature demands understanding of what variables mean, how they causally interact, and what transformations are semantically admissible. This semantic knowledge lies beyond what fixed operator libraries can encode.

Early automated feature engineering (AutoFE) systems frame the problem as search over a predefined transformation grammar~\citep{olson2016tpot, horn2019autofeat, zhang2023openfe}. While tractable, this formulation is fundamentally bounded by the expressiveness of the operator library: features requiring cross-variable semantic reasoning lie outside the reachable space by construction. The emergence of large language models (LLMs) introduced a qualitatively different capability: by pretraining on massive corpora of scientific, mathematical, and domain-specific text, LLMs can hypothesize meaningful transformations from column names, task descriptions, and background knowledge, effectively shifting the paradigm from fixed operator selection to open-ended feature program synthesis~\citep{hollmann2023large, han2024large}. However, existing LLM-based methods share a critical structural flaw: proposals are generated from static prompts with no memory of prior evaluations, leading to repeated proposals, no adaptation to dataset-specific patterns, and no mechanism to synthesize features requiring operators from multiple transformation families simultaneously.

Recent effort of combining LLMs with evolutionary search partially mitigates this limitation by maintaining an experience buffer of high-scoring programs as in-context demonstrations~\citep{abhyankar2025llm, gong2025evolutionary}. Yet even these methods operate over a single undifferentiated population. The feature program space is not homogeneous, it comprises qualitatively distinct transformation families (arithmetic interactions, statistical aggregates, temporal dynamics, relational encodings, nonlinear univariate maps), each encoding a different inductive bias about how predictive signal arises. Single-population search ignores this structure. Once a few successful programs from a dominant family saturate the experience buffer, subsequent LLM proposals become progressively confined to that family. Orthogonal families are under-explored, and cross-family compositions, which often yield the most expressive and complementary features, are structurally unreachable. This \emph{cross-family lock-in} is the central failure mode of existing LLM-evolutionary methods, and no existing work provides a principled mechanism to detect or correct it. Three interrelated limitations persist across the literature. \textit{(L1) Homogeneous search dynamics:} single-population evolution collapses onto a narrow subset of transformation types, biased toward early high-performing motifs and blind to the multi-modal structure of program space. \textit{(L2) Stateless LLM querying:} feature generation lacks persistent memory of prior exploration, preventing cumulative adaptation to dataset-specific patterns. \textit{(L3) Rigid transfer structure:} when multiple populations are used, migration is periodic or random, agnostic to when transfer is beneficial and which source family is most complementary to a stagnating target island.

We introduce \topo, a framework that models feature engineering as program search inspired by graph-structured multi-island evolution. \topo rests on three principles. First, \emph{inductive bias decomposition}: the program space is partitioned into different transformation families, each assigned a dedicated island that evolves specialized programs under family-specific priors, preserving diversity across transformation types by construction. Second, \emph{adaptive topology learning}: instead of fixed or heuristic migration schedules, \topo maintains a directed weighted graph over islands whose edge weights are updated online from observed cross-family transfer gains, learning a data-driven model of inter-family complementarity. Third, \emph{semantic transfer via LLM synthesis}: cross-island knowledge transfer is not realized by copying programs, but through LLM-mediated hybrid synthesis in which programs from a source and target island are jointly presented as context, and the LLM generates a compositionally novel program that inherits structural elements from both families. Transfer is triggered by a \emph{saturation criterion}: a sliding-window estimate of each island's marginal improvement that detects when local search has genuinely exhausted its productive region. This decouples transfer from wall-clock generation count and ensures that the evaluation budget is directed toward cross-family exploration precisely when it is most valuable. A \emph{Prompt Adaptation Memory} component further maintains a per-island accept/reject history and a compact natural-language summary of preferred and avoided transformation patterns, injected into every LLM call. This enables cumulative, dataset-specific adaptation throughout the search without any parameter update.

We summarize our contributions as follows: \textbf{(i) Topology-aware formulation.} We formulate AutoFE as graph-structured multi-island program search, partitioning the feature space into family-specialized islands and replacing heuristic migration with a learned directed topology graph whose edge weights encode empirically observed cross-family transfer utility, updated online via a bandit-style rule. \textbf{(ii) Saturation-triggered adaptive transfer.} We introduce a utility-based saturation criterion that triggers cross-island transfer when local search is exhausted. A per-island Prompt Adaptation Memory accumulates accepted and rejected program patterns, enabling cumulative dataset-specific adaptation throughout search without any parameter update. \textbf{(iii) Principled evaluation framework.} We introduce three purpose-built metrics: Mean Pairwise Output Correlation and Effective Rank jointly characterise feature redundancy and subspace coverage, and Topology Graph Specialisation Score quantifies the degree to which the topology graph acquires task-specific cross-family transfer knowledge during search. \textbf{(iv) Empirical validation.}  Across 29 datasets we demonstrate that \topo consistently outperforms most classical and LLM-based AutoFE baselines, producing lower-redundancy and higher-coverage feature sets, with performance remaining stable across downstream predictors and LLM backbones of varying capability.

\section{Preliminary}
\label{sec:method}
\subsection{Feature Engineering (FE)}
\label{sec:fe}
Let $\mathcal{D} = \{(\mathbf{x}_i, y_i)\}_{i=1}^{N}$ be a tabular dataset with $N$ instances, $d$-dimensional feature vector $\mathbf{x}_i \in \mathcal{X} \subseteq \mathbb{R}^{d}$ and target variable $y_i \in \mathcal{Y}$. Each dataset is accompanied by structured metadata $\mathcal{M} = \{\mathcal{M}^{\mathrm{task}}, \mathcal{M}^{\mathrm{feat}}\}$, encoding task type and per-feature semantic descriptors. Given a feature transformation
$\mathcal{T}: \mathcal{X} \to \tilde{\mathcal{X}}$, let $\mathcal{F}$ denote the class of downstream predictive models and $\mathcal{L}$ a task-appropriate loss function, the optimal downstream model is obtained via empirical risk minimization on training split $\{\bX_{tr}, {\by_{tr}}\}$:
\begin{equation}
  f_\mathcal{T}^* = \argmin_{f \in \cF}\;
  \cL\!\left(f\!\left(\mathcal{T}(\bX_{\mathrm{tr}})\right),\, \by_{\mathrm{tr}}\right).
  \label{eq:erm}
\end{equation}
The \emph{feature engineering objective} is to find the transformation
$\mathcal{T}^*$ that maximizes generalization performance on held-out data:
\begin{equation}
  \mathcal{T}^* = \argmax_{\mathcal{T}}\;
  \E\!\left[\Perf\!\left(f_\mathcal{T}^*,\,
  \mathcal{T}(\bX_{\mathrm{val}}),\, \by_{\mathrm{val}}\right)\right],
  \label{eq:feobjective}
\end{equation}
where $\Perf$ is a task-appropriate metric (e.g., accuracy). This constitutes a \emph{bilevel optimization}: the inner problem fits the model under a fixed transformation, while the outer searches for the transformation itself. As the outer objective is non-differentiable with respect to $\mathcal{T}$ and each
evaluation requires re-fitting $f$ from scratch, sample-efficient search strategies are essential.

\subsection{AutoFE as LLM-guided Program Search}
\label{sec:feat_search_llm}
We instantiate the transformation space $\mathcal{T}$ as a space of executable feature programs over a finite typed \emph{operator library} $\mathcal{O} = \{o_q\}_{q=1}^{Q}$, where each operator $o_q$ has a specified arity and type signature (e.g., $+, -, \times, \div, \log, \exp $). A \emph{feature program}
$p \in \mathcal{P}$ is any function $p: \mathcal{X} \to \mathbb{R}$ obtained by
composing operators from $\mathcal{O}$ up to a bounded depth $L_{\max}$:
\begin{equation}
\mathcal{P}
\coloneqq
\Bigl\{
p \;\Big|\;
p = o^{(L)} \circ \cdots \circ o^{(1)},
\; o^{(\ell)} \in \mathcal{O},
\; L \leq L_{\max}
\Bigr\}.
\end{equation}
Each $p \in \mathcal{P}$ is represented as an executable Python function. A feature set
$S \subseteq \mathcal{P}$ induces a transformed design matrix
$\mathcal{T}_{S}(\mathbf{X}) = [p(\mathbf{X})]_{p \in S}
\in \mathbb{R}^{N \times |S|}$, where each column corresponds to the output of one feature program. The optimal downstream model is:
\begin{equation}
f_{S}^{*}
=
\argmin_{f \in \mathcal{F}}
\mathcal{L}
\left(
f\left(\mathcal{T}_{S}(\mathbf{X}_{\mathrm{tr}})\right),
\mathbf{y}_{\mathrm{tr}}
\right).
\end{equation}
The feature engineering objective can therefore be formulated as a
\emph{combinatorial program search} problem:
\begin{equation}
S^{*}
=
\argmax_{S \subseteq \mathcal{P}}
\;
\mathbb{E}
\left[
\mathrm{Perf}
\left(
f_{S}^{*},
\mathcal{T}_{S}(\mathbf{X}_{\mathrm{val}}),
\mathbf{y}_{\mathrm{val}}
\right)
\right].
\label{eq:progsearch}
\end{equation} This search is intractable by enumeration because $|\mathcal{P}|$ grows
super-exponentially with $L_{\max}$, and non-differentiable with respect to discrete program structure. Each evaluation incurs cost 
$\mathcal{O}\!\left(k_{\mathrm{cv}} \cdot C_{\mathrm{fit}}(N, |S|)\right)$ under $k_{cv}$-fold cross-validation, where $C_{\mathrm{fit}}(N, |S|)$ denotes the training cost on $N$ instances and $|S|$ engineered features. The fitness signal $\hat{\Phi}(S)$ is estimated as:
\begin{equation}
\hat{\Phi}(S)
=
\frac{1}{k_{\mathrm{cv}}}
\sum_{k=1}^{k_{\mathrm{cv}}}
\mathrm{Perf}
\left(
f_{S,k}^{*},
\mathcal{T}_{S}(\mathbf{X}_{k}^{\mathrm{val}}),
\mathbf{y}_{k}^{\mathrm{val}}
\right),
\label{eq:oracle}
\end{equation}
where $f_{S,k}^{*}$ is trained on the $k$-th training fold and evaluated on the corresponding held-out fold $\bX_{val}, \by_{val}$. The metadata $\mathcal{M}$ restricts $\mathcal{P}$ to semantically valid programs and provides a natural-language prior exploited directly by LLM-guided synthesis.

\section{\topo}
\label{sec:topology_method}
\begin{figure*}[htb!]
    \centering
    \includegraphics[width=0.95\textwidth]{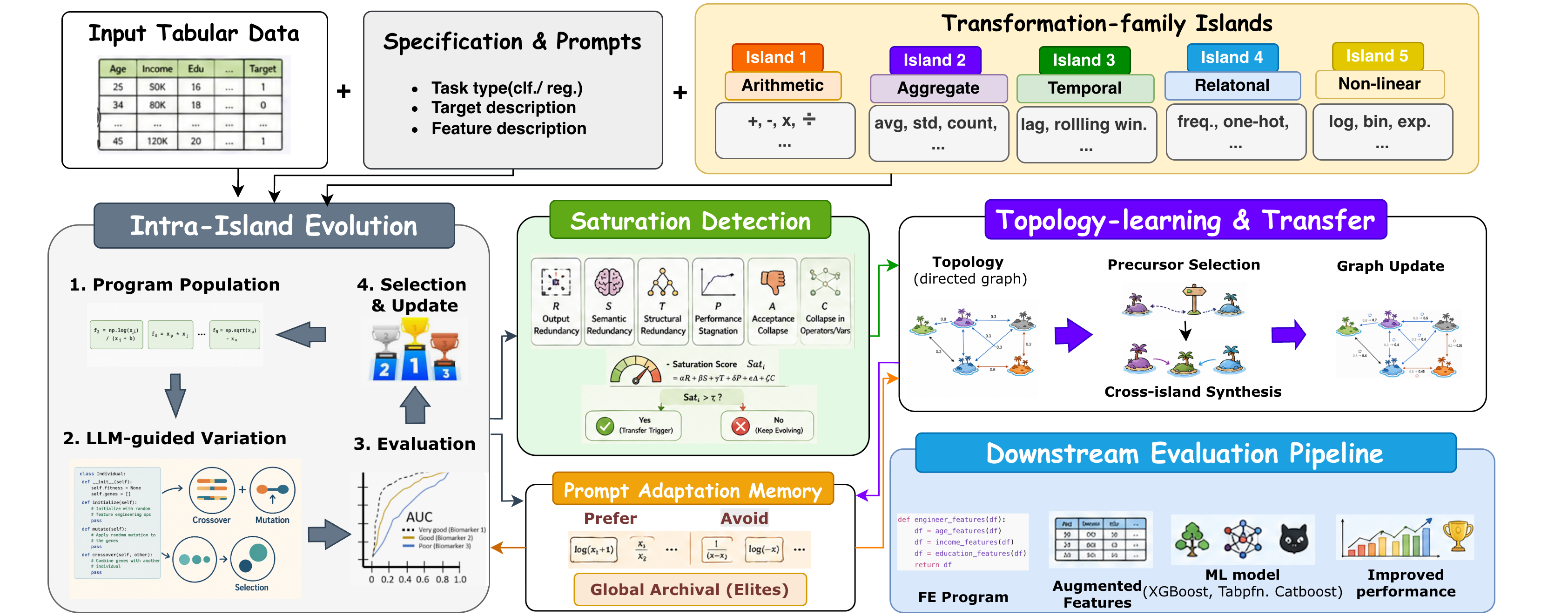}

    \caption{Overview of \topo. It decomposes FE-program search into multiple transformation-family islands with LLM-guided local evolution, adaptive saturation detection, topology-aware cross-island synthesis, and archive-based selection to generate diverse engineered features that improve downstream tabular prediction performance.}
    \label{fig:topofe}
\end{figure*}
Based on the preliminary stated in \S~\ref{sec:feat_search_llm}, we argue that, combinatorial program search over $\mathcal{P}$ is intractable by enumeration and inaccessible to gradient-based methods, requiring a LLM-guided proposal mechanism that satisfies three properties: (i) generate syntactically valid and semantically meaningful programs over a heterogeneous operator space without exhaustive enumeration, (ii) exploit task-specific domain knowledge in $\mathcal{M}$ to bias search toward promising and admissible transformations, and (iii) support diverse proposal modes within a unified interface, enabling both exploitation of known high-quality programs and exploration of unvisited regions of $\mathcal{P}$. To satisfy these requirements, we propose \topo, a framework that organizes LLM-guided program search for AutoFE through three coordinated components. First, a \emph{multi-island decomposition} partitions $\mathcal{P}$ into $M$ semantically coherent transformation families, each explored by a dedicated island to prevent premature convergence to a single family (\S\ref{sec:intraisland}). Second, \emph{Prompt Adaptation Memory} maintains per-island search state that continuously adapts LLM proposals from accumulated accept/reject history without any parameter update (\S\ref{sec:pam}). Third, a \emph{topology-aware cross-island transfer} mechanism governs knowledge flow across islands via a learned directed graph, triggering hybrid synthesis from complementary families when local search saturates (\S\ref{sec:transfer}). Together, these components balance specialized exploitation within transformation families and structured exploration across them. The overview of \topo is shown in Figure~\ref{fig:topofe} and the complete procedure is summarized in Algorithm~\ref{alg:topofe}.

\subsection{Intra-island Evolution}
\label{sec:intraisland}
\subsubsection{Multi-island Decomposition}
\label{sec:multiisland_decomp}
The feature program space $\mathcal{P}$ is not homogeneous: programs naturally cluster into semantically coherent regions according to the class of operations they employ, each encoding a distinct inductive bias about how predictive signal is constructed from raw features.
\begin{definition}[Transformation Family]
\label{def:family}
A \emph{transformation family} $\cP_i \subseteq \cP$ is a semantically coherent subset of feature programs sharing the same \emph{inductive bias} about how predictive signal is constructed from raw features, i.e., programs that apply the same class of primitive operations. \end{definition}

We consider five canonical transformation families: arithmetic interactions $\mathcal{P}_1$ 
(products, ratios, and differences), statistical aggregates $\mathcal{P}_2$ 
(group-level means, standard deviations, counts, and medians), temporal features 
$\mathcal{P}_3$ (lags, rolling windows, and exponential moving averages), relational 
encodings $\mathcal{P}_4$ (frequency counts, target encodings, and rank-based transforms), 
and nonlinear univariate maps $\mathcal{P}_5$ (logarithmic, exponential, binned, and 
polynomial transformations). These families may overlap (i.e., 
$\mathcal{P}_i \cap \mathcal{P}_j \neq \varnothing$ in general) but each represents a 
structurally distinct region of $\mathcal{P}$. The partition is also extensible to additional transformation families as needed. Together, they induce an approximate cover $\mathcal{P} \approx \bigcup_{i=1}^{M} \mathcal{P}_i$, which decomposes the monolithic search space into $M$ semantically typed 
subspaces, each amenable to specialized search.
 
However, in a single evolutionary population, high-scoring programs from a dominant family rapidly saturate the experience buffer, confining subsequent LLM proposals to that family's neighbourhood and leaving orthogonal families permanently unexplored. \topo addresses this through a structured multi-island decomposition that assigns a dedicated subpopulation to each $\mathcal{P}_i$, preserving family-specific inductive biases and preventing any single family from monopolizing the search budget.
\begin{definition}[Island]
\label{def:island}
Island $i$ at generation $t$ is a tuple
$\mathcal{I}_i^{(t)} = (\Pi_i^{(t)}, \mathcal{A}_i, \mathcal{H}_i^{(t)},
\rho_i^{(t)})$, where: (i) $\Pi_i^{(t)} \subset \cP_i$ is the current population of at most $n_{\max}$ feature programs, ranked by fitness $\hat\Phi$,
(ii) $\mathcal{A}_i \subset \cP_i$ is a long-term archive of elite programs
discovered by island $i$, (iii) $\mathcal{H}_i^{(t)}$ is the accept/reject history used to adapt the mutation prompt, and (iv) $\rho_i^{(t)} \in \R^s$ is a prompt-memory vector encoding a compact``prefer/avoid" signal derived from $\mathcal{H}_i^{(t)}$.
\end{definition}

These four components maintain the island's complete search state: $\Pi_i^{(t)}$ drives local search, $\mathcal{A}_i^{(t)}$ accumulates the best discoveries and serves as the in-context demonstration pool for LLM proposals, $\mathcal{H}_i^{(t)}$ provides the raw signal for prompt adaptation, and $\rho_i^{(t)}$ translates that signal into natural-language guidance injected into every LLM call.  
\subsubsection{LLM-guided Intra-island Evolution (Specialized Exploration)}
At each generation $t$, island $i$ independently evolves its population via two LLM-guided operators, both conditioned on the island's metadata context $\cM$ and prompt memory $\rho_i^{(t)}$:
\begin{itemize}
  \item \textbf{Mutation} samples a single parent $p\sim \Pi_i^{(t)}$ with probability proportional to $\hat{\Phi}(p)$ and prompts the LLM to produce a modified program $p'\in \mathcal{P}_i$ that improves on $p$ while respecting the signals encoded in $\rho_i^{(t)}$.
\item \textbf{Crossover} samples two parent programs $(p_a, p_b) \sim \Pi_i^{(t)}$ and presents them as in-context demonstrations. The LLM is then prompted to synthesize a child program $p' \in \mathcal{P}_i$ that combines compositional elements from both parents.
\end{itemize}

Both operators pass through a unified validation and fitness pipeline: each candidate $p'$ is checked for syntactic validity and semantic admissibility under $\mathcal{M}$, then evaluated via oracle $\hat{\Phi}$. The population $\Pi_i^{(t+1)}$ retains the top-$n_{\max}$ programs by fitness, and $\mathcal{A}_{i}^{(t)}$ and $\rho_i^{(t)}$ are updated following the rules defined in~\S\ref{sec:pam}.

\subsubsection{Global Objective}
The multi-island system assembles the final feature set by pooling elites across all islands:
\begin{equation}
  S^* = \argmax_{S \,\subseteq\, \bigcup_i \mathcal{A}_i}
  \hat\Phi(S),
  \quad
  \text{s.t.}\ |S| \leq d'_{\max},
  \quad
  \sum_{i=1}^{K} |\Pi_i^{(t)}| \leq B,
  \label{eq:globalobj}
\end{equation}
where $B$ is the total evaluation budget and each retained program must contribute non-redundant predictive signal:
 $\hat\Phi(\{p\} \mid S^* \setminus \{p\}) > 0\ \forall p \in
S^*$.
\subsection{Prompt Adaptation Memory}
\label{sec:pam}
While the LLM's pretrained knowledge provides a strong prior over feature programs, effective search requires the proposal mechanism to \emph{adapt online} to each island's evolving landscape. Without adaptation, the LLM repeatedly proposes programs in regions already known to be unproductive, wasting evaluation budget. We introduce \emph{Prompt Adaptation Memory} (PAM), a lightweight per-island mechanism that continuously shapes the LLM's proposal distribution from accumulated search experience without any parameter update. PAM operates through three coordinated update rules applied after every generation.

\subsubsection{Prompt-memory Update} Let $\mathcal{H}_i^{+}(t)$ and $\mathcal{H}_i^{-}(t)$ denote the accepted and rejected subsets of the sliding history window $[t-W, t]$. Operator types concentrated among accepted programs are reinforced as \textit{prefer} signals, those concentrated among rejected programs are flagged as \textit{avoid} signals. Formally:
\begin{equation}
\rho_i^{(t+1)} =
\mathrm{Summarize}\!\left(
\mathcal{H}_i^{+}(t),
\mathcal{H}_i^{-}(t),
\rho_i^{(t)}
\right),
\label{eq:pamupdate}
\end{equation}
where $\mathrm{Summarize}(\cdot)$ is an LLM call that produces a compact natural-language string (e.g., ``prefer ratio features with lagged denominators, avoid log transforms of sparse columns") prepended to all subsequent prompts for island $i$.

\subsubsection{Elite Archive Update} After each generation, $\mathcal{A}_i^{(t)}$ merges newly accepted programs, re-ranks by $\hat{\Phi}$, and truncating to capacity $|\mathcal{A}|_{\max}$:
\begin{equation}
\mathcal{A}_i^{(t+1)} =
\mathrm{TopK}\!\left(
\mathcal{A}_i^{(t)} \cup \{p : a_t = 1\},
|\mathcal{A}|_{\max},
\hat{\Phi}
\right).
\label{eq:archiveupdate}
\end{equation}

Candidates whose normalized tree-edit distance to any existing archive member falls below $\delta_{\min}$ are discarded before insertion, enforcing structural novelty. The archive serves as both the in-context demonstration pool for intra-island proposals and the candidate source for cross-island synthesis.

\subsubsection{Proposal Conditioning}
Every LLM call for island $i$ at generation $t$ is conditioned on:
\begin{equation}
\mathrm{Context}_i^{(t)} =
\left(
\mathcal{M},\,
\rho_i^{(t)},\,
\mathrm{Sample}\!\left(\mathcal{A}_i^{(t)}, n\right)
\right),
\label{eq:context}
\end{equation}
where $\mathcal{M}$ grounds the proposal in domain knowledge, $\rho_i^{(t)}$steers it with accumulated search preferences, and $\mathrm{Sample}(\mathcal{A}_i^{(t)}, n)$ exemplifies it with the $m$ highest-scoring programs discovered so far. This three-way conditioning closes the feedback loop introduced above: history shapes memory, memory steers proposals, and accepted proposals enrich the archive.
\subsection{Topology-Aware Cross-Island Transfer}
\label{sec:transfer}
Multi-island decomposition prevents within-family saturation but cannot discover features requiring operators from multiple families simultaneously. For instance, the feature
$\frac{\mathrm{rolling\_mean}(\texttt{sales},7)}{\mathrm{lag}(\texttt{sales},3)}$
combines temporal operators ($\cP_3$) with arithmetic operators ($\cP_1$)
neither family alone can generate it. We introduce topology-aware cross-island transfer to systematically discover such cross-family compositions, organized around three questions: \emph{when} to transfer (saturation detection), \emph{where} to transfer (precursor selection), and \emph{what} to transfer (hybrid synthesis).

\begin{definition}[Topology Graph]
\label{def:topograph}
The \emph{topology graph} $\cG^{(t)} = (V, E, \mathbf{W}^{(t)})$ is a directed weighted graph where $V$ is the set of islands, $E$ is the set of directed transfer edges, and $w_{j \to i}^{(t)}\in \mathbf{W}^{(t)}$ encodes the empirically observed utility of transferring knowledge from island $j$ to $i$ up to generation $t$. \end{definition} 

Edge weights are initialized uniformly, $w_{j \to i}^{(0)} = 1/(M-1)$ for $j \neq i$ and updated online after each transfer event via an exponential moving average:
\begin{equation}
  w_{j \to i}^{(t+1)} =
  (1-\alpha)\,w_{j \to i}^{(t)}
  + \alpha\,\Delta_{\mathrm{cross}}^{(j \to i)}(t),
  \label{eq:weightupdate}
\end{equation}
where $\alpha \in (0,1)$ is a decay coefficient and
$\Delta_{\mathrm{cross}}^{(j \to i)}(t)$ is observed fitness gain from the transfer. This makes $\cG^{(t)}$ a continuously refined, data-driven model of cross-family complementarity updated online throughout the search.

\subsubsection{When to Transfer: Saturation Detection.} An island is saturated when it has exhausted the useful information extractable under its current transformation family, so that further local search yields diminishing returns. Formally, island $i$ is \emph{saturated} at generation $t$ if its expected marginal improvement from intra-island proposals falls below threshold  $\varepsilon > 0$ over a sliding window of $W$ consecutive generations:
\begin{equation}
  \frac{1}{W}\sum_{\tau=t-W+1}^{t}
  \Delta_i(\tau) \leq \varepsilon,
  \label{eq:saturation}
\end{equation}
where $\Delta_i(\tau) = \E\!\left[U_i^{\max}(\tau+1) - U_i^{\max}(\tau)\right]$, 
and $U_i^{\max}(\tau) = \max_{p \in \Pi_i^{(\tau)}} \hat\Phi(p)$. Saturation detection decouples transfer from fixed schedules, triggering cross-island communication only when local search has genuinely stagnated.

\subsubsection{Where to Transfer: Precursor Selection} When island $i$ is saturated, \topo selects a \emph{precursor island} $j^*$, the donor whose knowledge is most likely
to alleviate $i$'s stagnation via:
\begin{equation}
  j^* = \argmax_{j \neq i}\; w_{j \to i}^{(t)}
  \label{eq:precursor}
\end{equation} 
with ties broken at random. Over time, $\cG^{(t)}$ identifies persistently high-utility transfer pathways (e.g., temporal
$\to$ arithmetic), concentrating the search budget on productive cross-family compositions.
\subsubsection{What to Transfer: Cross-Island Synthesis} Given saturated island $i$ and precursor $j^*$, cross-island synthesis
generates hybrid programs in the joint composition space:
\begin{equation}
  \cP_{i \leftarrow j^*} \subseteq
  \cP_i \cup \bigl(\cP_i \circ \cP_{j^*}\bigr)
  \label{eq:compspace}
\end{equation}
where $\circ$ denotes operator-level composition. The LLM is prompted with the top-$m$ programs from $\Pi_i^{(t)}$ as target context, the top-$m$ programs from $\Pi_{j^*}^{(t)}$ as donor context, and the task metadata $\cM$, producing programs that integrate structural elements from $\mathcal{P}_{j^{*}}$, while remaining consistent with $\mathcal{P}_i$'s inductive bias. This \emph{LLM-mediated hybrid synthesis} differs fundamentally from direct program copying, generating programs unreachable by intra-island search alone. Accepted candidates update $\Pi_i^{(t)}$ and $\mathcal{A}_i^{(t)}$, the observed gain then updates $w_{j^* \to i}^{(t)}$ via Eq.(~\ref{eq:weightupdate}), ensuring future routing decisions reflect the empirical utility of this transfer pathway.

\subsubsection{Unified Objective}

The full \topo objective jointly optimizes predictive performance, feature complementarity, stability, and computational efficiency. Let $\mathcal{A} = \bigcup_{i=1}^{M} \mathcal{A}_i$
denote the pooled candidate archive assembled from all island archives. The final engineered feature set $S^* \subseteq \mathcal{A}$ is obtained through redundancy-aware selection under a cardinality constraint $|S^*| \le d'_{\max}$. The optimization objective is defined as:
\begin{align}
\max_{\{\Pi_i\},\, \mathcal{G},\, S^* \subseteq \mathcal{A}}
\quad &
\E\!\left[
\Perf\!\left(
f_{S^*}^*,
T_{S^*}(\bX_{\mathrm{val}}),
\by_{\mathrm{val}}
\right)
\right]
\notag\\
&
-\lambda_1\,\mathrm{Redundancy}(S^*)
+\lambda_2\,\mathrm{Stability}(S^*)
\notag\\
&
-\lambda_3\,\mathrm{Cost}(B)
\label{eq:fullobj}
\\[0.3em]
\text{s.t.}\quad
&
|S^*| \le d'_{\max}.
\notag
\end{align}

Here, the predictive term evaluates the downstream generalization performance of the model trained on the engineered feature set $S^*$. The redundancy penalty discourages highly correlated feature programs: $\mathrm{Redundancy}(S)
=
\frac{1}{|S|(|S|-1)}
\sum_{p \neq p' \in S}
\left|
\mathrm{Corr}\!\left(
p(\bX),\, p'(\bX)
\right)
\right|$, where $\mathrm{Corr}(\cdot,\cdot)$ denotes Spearman rank correlation computed over feature outputs. This formulation excludes self-correlations and penalizes only pairwise redundancy between distinct feature programs. The stability term
$\mathrm{Stability}(S)$ measures the inverse variance of the fitness estimate $\hat{\Phi}(S)$ across cross-validation folds, encouraging feature sets whose performance remains consistent under different train-validation partitions. The computational term $\mathrm{Cost}(B)$ represents the total oracle evaluation budget consumed during search. The hyperparameters $\lambda_1,\lambda_2,\lambda_3 \geq 0$ control the trade-off between predictive accuracy, feature diversity, robustness, and computational efficiency.

After search termination, the final feature set is constructed by greedily selecting high-scoring programs from $\mathcal{A}$ in descending validation-performance order while discarding candidates whose absolute correlation with any previously selected feature exceeds a predefined threshold $\tau_{\mathrm{red}}$.
\iffalse
\subsubsection{Unified Objective} The full \topo  objective jointly optimizes predictive performance, feature complementarity, stability, and computational cost:
\begin{align}
  \max_{\{\Pi_i\}, \cG} \quad &
  \E\!\left[\Perf(f_{S^*}^*,\, T_{S^*}(\bX_{\mathrm{val}}),\, \by_{\mathrm{val}})\right]
  \notag\\
  &- \lambda_1\,\mathrm{Redundancy}(S^*)
  + \lambda_2\,\mathrm{Stability}(S^*)
  \notag\\
  &- \lambda_3\,\mathrm{Cost}(B)
  \label{eq:fullobj}
\end{align}
where $S^* = \bigcup_i \mathcal{A}_i$ is the union of island archives. $\mathrm{Redundancy}(S) = \frac{1}{|S||S-1|}\sum_{p,p' \in S, p\ne p'}
|\mathrm{Corr}(p(\bX), p'(\bX))|$ penalizes collinear features. \ 
$\mathrm{Stability}(S)$ measures the variance of $\hat\Phi(S)$ across
cross-validation folds. $\mathrm{Cost}(B)$ is the total number of oracle evaluations consumed. The hyperparameters $\lambda_1,\lambda_2,\lambda_3 \geq 0$ balance these objectives.
\fi
\begin{algorithm}[t]
\caption{\topo}
\label{alg:topofe}
\small
\KwIn{$\mathcal{D}$, $\mathcal{M}$, $\{\mathcal{P}_i\}_{i=1}^{M}$, budget $B$, window $W$, threshold $\varepsilon$, decay $\alpha$}
\KwOut{$S^{*}$}

\For{$i=1,\ldots,K$}{
    $\Pi_i^{(0)} \leftarrow \mathrm{LLM\text{-}Seed}(\mathcal{P}_i,\mathcal{M})$;
    $\mathcal{A}_i^{(0)} \leftarrow \varnothing$;
    $\rho_i^{(0)} \leftarrow \varnothing$\;
}
Initialize $\mathcal{G}^{(0)}$ with $w_{j\to i}^{(0)}=1/(M-1)$ for all $j\neq i$;
$t\leftarrow0$, $b\leftarrow0$\;

\While{$b<B$}{
    \ForPar{$i=1,\ldots,M$}{
        $p' \leftarrow
        \mathrm{LLM\text{-}Mutate/Crossover}(\mathcal{A}_i^{(t)},\mathcal{M},\rho_i^{(t)})$\;
        Evaluate $\hat{\Phi}(p')$; $b\leftarrow b+1$\;
        Update $\Pi_i^{(t+1)},\mathcal{A}_i^{(t+1)},\mathcal{H}_i^{(t+1)},\rho_i^{(t+1)}$
        via Eqs.~\ref{eq:pamupdate},~\ref{eq:archiveupdate}\;
    }

    \For{$i=1,\ldots,M$}{
        \If{$\mathrm{Saturated}(i,W,\varepsilon)$}{
            $j^* \leftarrow \argmax_{j\neq i} w_{j\to i}^{(t)}$\;
            $p' \leftarrow
            \mathrm{LLM\text{-}HybridSynth}(\mathcal{A}_i^{(t)},\mathcal{A}_{j^*}^{(t)},\mathcal{M})$\;
            Evaluate $\hat{\Phi}(p')$; $b\leftarrow b+1$\;
            Update $\Pi_i^{(t+1)},\mathcal{A}_i^{(t+1)}$\;
            $w_{j^*\to i}^{(t+1)}
            \leftarrow
            (1-\alpha)w_{j^*\to i}^{(t)}
            +\alpha\Delta_{\mathrm{cross}}^{(j^*\to i)}(t)$\;
        }
    }
    $t\leftarrow t+1$\;
}
\Return{$S^* \leftarrow
\mathrm{GreedySelect}\!\left(\bigcup_{i=1}^{M}\mathcal{A}_i^{(t)},d'_{\max}\right)$}
\end{algorithm}
 
\subsection{Computational Complexity}
\label{sec:complexity}
Standard single-population evolutionary feature search evaluates $n$ candidates per generation, each at cost $C_{eval}$, giving a total cost of $\mathcal{O}(T \cdot n \cdot C_{\mathrm{eval}})$ over $T$ generations. \topo distributes $n$ candidates across $M$ islands with $\sum_i n_i = n$, incurring per-generation cost:
\begin{equation}
\mathcal{O}\!\left(\sum_{i=1}^M n_i \cdot C_{\mathrm{eval}} + |\mathcal{E}_t| \cdot C_{\mathrm{transfer}}\right)
\label{eq:cost}
\end{equation} where $|\mathcal{E}_t|$ is the number of active transfer edges and $\mathrm{transfer}$ is the cost of a single hybrid synthesis call. Under balanced allocation ($n_i \approx n/M$), this reduces to $\mathcal{O}(n \cdot C_{\mathrm{eval}} + |\mathcal{E}_t| \cdot C_{\mathrm{transfer}})$, matching the single-population baseline up to transfer overhead. Since transfer is saturation-triggered rather than applied at every generation, $|\mathcal{E}_t|$ remains small in practice and transfer overhead is negligible.
\paragraph{Parallelism.} Island evaluations are mutually independent, so all $M$ islands execute in parallel, reducing wall-clock time per generation from $\mathcal{O}(n \cdot C_{\mathrm{eval}})$ to $\mathcal{O}(\max_i n_i \cdot C_{\mathrm{eval}})$ and yielding an approximate $k$-fold speedup under balanced partitioning.
\paragraph{Sample Efficiency.} Saturation-triggered transfer concentrates oracle evaluations on high-utility regions of $\mathcal{P}$, reducing the number of generations $T$ required to converge relative to single-population search. \topo thus achieves comparable asymptotic complexity to the baseline while improving both wall-clock efficiency through parallelism and convergence speed through principled cross-family exploration.

\section{Experiments}
\label{sec:experiments}
We conduct experiments to address the following research questions. \textbf{RQ1:} Can \topo outperform existing state-of-the-art AutoFE methods across diverse tabular learning tasks and dataset characteristics? (\S\ref{sec:main_res})  \textbf{RQ2:} Is \topo's performance robust to the choice of LLM backbone across models of varying capability? (\S\ref{sec:generalize}) \textbf{RQ3:} Are the feature programs discovered by \topo transferable across architecturally distinct downstream predictors beyond the search-time evaluation model? (\S\ref{sec:feat_trans}) \textbf{RQ4:} Does \topo produce feature sets that are more diverse, low-redundancy, and informationally complementary than competing methods? (\S\ref{sec:feature_disversity}) \textbf{RQ5:} Does the adaptive topology graph acquire meaningful task-specific knowledge about cross-family transfer utility during search? (\S\ref{sec:learning_effec}) \textbf{RQ6:} How each component of \topo contributes to the its overall performance? (\S\ref{sec:ablation}) 
\paragraph{Datasets}
We evaluate \topo on 29 public tabular datasets comprising 19 classification and 10 regression tasks, sourced from the UCI Machine Learning Repository~\citep{asuncion2007uci}\footnote{\url{https://archive.ics.uci.edu/datasets}}, Kaggle\footnote{\url{https://www.kaggle.com/datasets}}, and OpenML~\citep{vanschoren2014openml} \footnote{\url{https://www.openml.org/}}, following the dataset selection protocol of prior work~\citep{hollmann2022tabpfn, han2024large, abhyankar2025llm}. The benchmark spans diverse prediction tasks, dataset scales ($n_{inst}\in[315, 581012]$), feature dimensionalities ($n_{feat} \in [4, 279]$), and heterogeneous feature types including numerical, categorical, and temporal variables. Dataset statistics are reported in Tables~\ref{tab:classification_main} and~\ref{tab:regression_main}. Each dataset is accompanied by structured metadata $\mathcal{M}$ comprising task-level descriptions and per-feature semantic annotations used to condition LLM generation. All datasets are partitioned using a fixed $80/20$ train/test split, where the training partition is used exclusively for feature search, cross-validation, and model fitting, and the test partition is held out for final unbiased evaluation. All experiments are repeated over five runs with different random seeds and data splits. Results are reported as mean $\pm$ standard deviation.

\paragraph{Baselines} 
We compare \topo against six SOTA AutoFE methods spanning three categories: (i) \emph{classical methods}: OpenFE~\citep{zhang2023openfe} and AutoFeat~\citep{horn2019autofeat}, (ii) \emph{LLM-based methods}: CAAFE~\citep{hollmann2023large}, FeatLLM~\citep{han2024large}, and OCTree~\citep{nam2024optimized}, and (iii) \emph{LLM-guided evolutionary search}: LLM-FE~\citep{abhyankar2025llm}, the most closely related baseline sharing the multi-population evolutionary architecture. All LLM-based methods use the same backbone model and generation settings as \topo to ensure a controlled comparison.
 
\paragraph{Evaluation Protocol} Classification performance is measured by accuracy ($\uparrow$) and regression performance by Root Mean Square Error ($\downarrow$). Feature evaluation follows a two-stage procedure: given a candidate feature set $S$, we first execute each program $p\in S$ on the training data to construct the augmented representation $\mathcal{T}_S(\mathbf{X}_{\mathrm{tr}})$, then train a downstream predictor $f_S^*$ on the augmented training set and evaluate the fitness signal $\hat{\Phi}(S)$ on the held-out validation fold via 5-fold stratified cross-validation as defined in Eq.~(\ref{eq:oracle}). For fair comparison across all methods, XGBoost is used as the default downstream predictor with identical train/validation protocols, evaluation budgets, and model configurations unless explicitly stated otherwise.
\paragraph{Implementation Details.}
All LLM-based methods are implemented with three backbone models: Qwen3-8B~\citep{yang2025qwen3}, Qwen2.5-Coder~\citep{hui2024qwen2}, and GPT-4o-mini~\citep{hurst2024gpt}. Unless otherwise specified, we use a sampling temperature of $\tau=0.8$ for all LLM genertion calls. For \topo, we instantiate $M=5$ family-specialised islands corresponding to the five canonical transformation families defined in \S\ref{sec:intraisland}: arithmetic interactions ($\mathcal{P}_1$), statistical aggregates ($\mathcal{P}_2$), temporal features ($\mathcal{P}_3$), relational encodings ($\mathcal{P}_4$), and nonlinear univariate maps ($\mathcal{P}_5$). At each generation, each LLM call produces $b = 3$ candidate feature programs written as executable Python functions.

Each LLM prompt comprises four components: (i) task-level metadata $\mathcal{M}^{task}$ summarizing the prediction objective and target variable, (ii) feature-level metadata $\mathcal{M}^{feat}$ comprising feature names, value types, and one-sentence semantic descriptions, with representative category values included for categorical variables, (iii) in-context demonstrations drawn from the island's elite archive $\mathcal{A}_{i}^{(t)}$, providing high-quality previously discovered programs as exemplars for mutation, crossover, or hybrid synthesis, and (iv) a strict output format specification requiring a syntactically valid Python feature-transformation function. A small number of serialized training instances with ground-truth labels are additionally included as task context, following~\citep{dinh2022lift, han2024large}. All generated programs undergo systematic validation comprising syntax checking, type consistency verification, numerical-safety filtering (e.g., division-by-zero and overflow guards), and metadata-consistency checks before fitness evaluation.

For the PAM component, the sliding history window is set to $W = 5$ generations and the prompt-memory string is updated after every generation via an LLM summarization call conditioned on $\mathcal{H}_i^+(t)$ and $\mathcal{H}_i^-(t)$. The elite archive capacity is $|\mathcal{A}|_{\max} = 20$ programs per island, with structural novelty enforced via a normalized tree-edit distance threshold of $\delta_{\min} = 0.1$. For the topology graph, the exponential moving average decay coefficient is $\alpha = 0.3$, and the saturation threshold is $\varepsilon = 0.01$ over a window of $W = 5$ consecutive generations. The redundancy penalty coefficient is $\lambda_1 = 0.3$, 
stability weight $\lambda_2 = 0.1$, and cost weight $\lambda_3 = 0.1$ in the unified objective (Eq.~\ref{eq:fullobj}). After search termination, the final feature set is assembled by pooling elite programs from all island archives and selecting at most $5$ non-redundant programs via greedy validation-score-ranked selection, discarding candidates whose absolute Spearman correlation with any already-selected feature exceeds $\tau_{\mathrm{red}} = 0.9$.

\section{Results and Analysis}
\label{sec:results}

\subsection{Main Results}
\label{sec:main_res}
\begin{table*}[!t]
\centering
\small
\caption{Performance of XGBoost on classification datasets with different AutoFE methods, measured by accuracy ($\uparrow$). Results are reported in mean $\pm$ std over five splits. Best in \textbf{bold}, second-best \underline{underlined}. Qwen3-8B is used as the backbone for all LLM-based methods. $n_{inst}$ is the number of instances and $n_{feat}$ is the number of features. `Base' represents the performance of XGBoost w/o FE.} 
\resizebox{\textwidth}{!}{
\begin{tabular}{lcccccccccccc}
\toprule
\multirow{2}{*}{Dataset} & \multirow{2}{*}{$n_{inst}$} & \multirow{2}{*}{$n_{feat}$} & \multirow{2}{*}{Base} 
& \multicolumn{2}{c}{Classical AutoFE} 
& \multicolumn{3}{c}{LLM-based FE}
& \multicolumn{2}{c}{LLM$+$Evolutionary FE} \\

\cmidrule(lr){5-6} \cmidrule(lr){7-9} \cmidrule(lr){10-11}

& & & 
& AutoFeat & OpenFE 
& CAAFE & FeatLLM & OCTree 
& LLMFE & \textbf{\topo} \\

\midrule
adult & 48842 & 14 &
0.8611 $\pm$ 0.0165 &
0.9265 $\pm$ 0.0011 &
\underline{0.9267} $\pm$ 0.0013 &
0.8784 $\pm$ 0.0052 &
0.8671 $\pm$ 0.0298 &
0.8668 $\pm$ 0.0104 &
0.8708 $\pm$ 0.0011 &
\textbf{0.9268} $\pm$ 0.0025 \\

arrhythmia & 452 & 279 &
0.7053 $\pm$ 0.0186 &
0.6702 $\pm$ 0.0371 &
\textbf{0.7344} $\pm$ 0.0362 &
0.7143 $\pm$ 0.0127 &
0.6561 $\pm$ 0.0296 &
0.6560 $\pm$ 0.0335 &
0.6566 $\pm$ 0.0129 &
\underline{0.7144} $\pm$ 0.0203 \\

balance-scale & 625 & 4 &
0.8320 $\pm$ 0.0233 &
0.8848 $\pm$ 0.0079 &
\underline{0.8560} $\pm$ 0.0036 &
0.8320 $\pm$ 0.0054 &
0.8252 $\pm$ 0.0315 &
0.7822 $\pm$ 0.0241 &
0.8780 $\pm$ 0.0093 &
\textbf{0.9465} $\pm$ 0.0046 \\

bank & 45211 & 16 &
0.8991 $\pm$ 0.0118 &
0.9020 $\pm$ 0.0019 &
\underline{0.9297} $\pm$ 0.0015 &
0.9091 $\pm$ 0.0031 &
0.9097 $\pm$ 0.0030 &
0.8985 $\pm$ 0.0228 &
0.9054 $\pm$ 0.0075 &
\textbf{0.9333} $\pm$ 0.0182 \\

breast-w & 699 & 9 &
0.9500 $\pm$ 0.0096 &
0.9728 $\pm$ 0.0054 &
0.9906 $\pm$ 0.0005 &
0.9429 $\pm$ 0.0043 &
\underline{0.9913} $\pm$ 0.0252 &
0.9597 $\pm$ 0.0130 &
0.9607 $\pm$ 0.0036 &
\textbf{0.9942} $\pm$ 0.0014 \\

blood & 748 & 4 &
0.6733 $\pm$ 0.0018 &
0.6805 $\pm$ 0.0056 &
0.6924 $\pm$ 0.0167 &
\underline{0.7733} $\pm$ 0.0100 &
\textbf{0.7749} $\pm$ 0.0051 &
0.7246 $\pm$ 0.0312 &
0.7208 $\pm$ 0.0209 &
0.7440 $\pm$ 0.0129 \\

car & 1728 & 6 &
0.9671 $\pm$ 0.0027 &
0.9872 $\pm$ 0.0030 &
\underline{0.9913} $\pm$ 0.0062 &
0.9819 $\pm$ 0.0080 &
0.8150 $\pm$ 0.0087 &
0.9783 $\pm$ 0.0164 &
0.9734 $\pm$ 0.0095 &
\textbf{0.9971} $\pm$ 0.0051 \\

diabetes & 253680 & 21 &
0.8491 $\pm$ 0.0068 &
0.8012 $\pm$ 0.0143 &
0.8483 $\pm$ 0.0005 &
0.8491 $\pm$ 0.0096 &
\underline{0.8497} $\pm$ 0.0040 &
0.8305 $\pm$ 0.0069 &
0.8298 $\pm$ 0.0054 &
\textbf{0.8524} $\pm$ 0.0021 \\

cmc & 1473 & 9 &
0.5051 $\pm$ 0.0075 &
0.5261 $\pm$ 0.0108 &
\underline{0.5301} $\pm$ 0.0112 &
0.5051 $\pm$ 0.0115 &
0.4800 $\pm$ 0.0020 &
0.5111 $\pm$ 0.0334 &
0.5170 $\pm$ 0.0194 &
\textbf{0.5577} $\pm$ 0.0078 \\

communities & 1994 & 103 &
0.6943 $\pm$ 0.0239 &
0.6991 $\pm$ 0.0075 &
0.6936 $\pm$ 0.0041 &
\underline{0.6992} $\pm$ 0.0079 &
0.5938 $\pm$ 0.0013 &
0.6917 $\pm$ 0.0110 &
0.6947 $\pm$ 0.0105 &
\textbf{0.7053} $\pm$ 0.0043 \\

covtype & 581012 & 54 &
0.8652 $\pm$ 0.0120 &
0.8668 $\pm$ 0.0153 &
0.8684 $\pm$ 0.0006 &
0.8652 $\pm$ 0.0076 &
0.8585 $\pm$ 0.0081 &
0.8332 $\pm$ 0.0115 &
\underline{0.8773} $\pm$ 0.0087 &
\textbf{0.8774} $\pm$ 0.0038 \\

credit-g & 1000 & 20 &
0.7000 $\pm$ 0.0237 &
0.7629 $\pm$ 0.0019 &
\underline{0.7725} $\pm$ 0.0221 &
0.7000 $\pm$ 0.0107 &
0.7124 $\pm$ 0.0076 &
0.7373 $\pm$ 0.0020 &
0.7500 $\pm$ 0.0147 &
\textbf{0.7878} $\pm$ 0.0066 \\

eucalyptus & 736 & 19 &
0.6724 $\pm$ 0.0151 &
0.6643 $\pm$ 0.0213 &
\underline{0.6766} $\pm$ 0.0180 &
0.6284 $\pm$ 0.0093 &
0.6191 $\pm$ 0.0062 &
0.6416 $\pm$ 0.0267 &
0.6514 $\pm$ 0.0080 &
\textbf{0.6990} $\pm$ 0.0094 \\

heart & 918 & 11 &
0.8696 $\pm$ 0.0190 &
0.8923 $\pm$ 0.0012 &
\underline{0.9199} $\pm$ 0.0043 &
0.8696 $\pm$ 0.0093 &
0.8789 $\pm$ 0.0160 &
0.8217 $\pm$ 0.0093 &
0.8352 $\pm$ 0.0163 &
\textbf{0.9346} $\pm$ 0.0015 \\

jungle-chess & 44819 & 6 &
0.8636 $\pm$ 0.0019 &
0.8446 $\pm$ 0.0202 &
\underline{0.8707} $\pm$ 0.0020 &
0.8636 $\pm$ 0.0088 &
0.5915 $\pm$ 0.0252 &
0.7796 $\pm$ 0.0096 &
0.8693 $\pm$ 0.0146 &
\textbf{0.8858} $\pm$ 0.0104 \\

myocardial & 686 & 92 &
0.7246 $\pm$ 0.0087 &
0.7232 $\pm$ 0.0138 &
0.7217 $\pm$ 0.0350 &
0.7536 $\pm$ 0.0114 &
0.7824 $\pm$ 0.0056 &
0.7827 $\pm$ 0.0175 &
\underline{0.7847} $\pm$ 0.0036 &
\textbf{0.7878} $\pm$ 0.0066 \\

pc1 & 1109 & 21 &
0.9214 $\pm$ 0.0135 &
0.9313 $\pm$ 0.0153 &
0.9367 $\pm$ 0.0254 &
0.9414 $\pm$ 0.0101 &
\textbf{0.9605} $\pm$ 0.0295 &
0.9302 $\pm$ 0.0186 &
0.9312 $\pm$ 0.0102 &
\underline{0.9456} $\pm$ 0.0066 \\

tic-tac-toe & 958 & 9 &
0.9896 $\pm$ 0.0022 &
0.9923 $\pm$ 0.0012 &
\textbf{0.9999} $\pm$ 0.0001 &
0.9896 $\pm$ 0.0134 &
0.6582 $\pm$ 0.0080 &
0.9886 $\pm$ 0.0058 &
0.9896 $\pm$ 0.0138 &
\underline{0.9993} $\pm$ 0.0220 \\

vehicle & 846 & 18 &
0.7617 $\pm$ 0.0019 &
0.7342 $\pm$ 0.0143 &
0.7541 $\pm$ 0.0067 &
\underline{0.7647} $\pm$ 0.0085 &
0.7519 $\pm$ 0.0106 &
0.7416 $\pm$ 0.0018 &
0.7574 $\pm$ 0.0033 &
\textbf{0.7781} $\pm$ 0.0073 \\

% continue remaining rows...
\bottomrule
\end{tabular}
}
\label{tab:classification_main}

\end{table*}

\begin{table*}[!t]
\centering
\small
\caption{Performance of XGBoost on regression datasets with different AutoFE methods, measured by RMSE $\downarrow$ (FeatLLM and CAAFE are designed specific for classification tasks only).}
\resizebox{\textwidth}{!}{
\begin{tabular}{lccccccccc}
\toprule
\multirow{2}{*}{Dataset} & \multirow{2}{*}{$n_{inst}$} & \multirow{2}{*}{$n_{feat}$} & \multirow{2}{*}{Base} 
& \multicolumn{2}{c}{Classical AutoFE} 
& \multicolumn{1}{c}{LLM-based FE}
& \multicolumn{2}{c}{LLM$+$Evolutionary FE} \\

\cmidrule(lr){5-6} \cmidrule(lr){7-7} \cmidrule(lr){8-9}

& & & 
& AutoFeat & OpenFE 
& OCTree 
& LLMFE & TopoFE \\

\midrule
airfoil\_self\_noise & 1503 & 6 &
1.6701 $\pm$ 0.0514 &
1.6265 $\pm$ 0.1701 &
1.6143 $\pm$ 0.0232 &
1.6701 $\pm$ 0.0763 &
\underline{1.5415} $\pm$ 0.0162 &
\textbf{1.3000 $\pm$ 0.0123} \\

bike & 17389 & 12 &
4.3494 $\pm$ 0.0745 &
4.4713 $\pm$ 0.1830 &
\underline{4.0471} $\pm$ 0.0576 &
4.2336 $\pm$ 0.1273 &
4.3494 $\pm$ 0.2363 &
\textbf{3.0210 $\pm$ 0.0202} \\

cpu & 8192 & 10 &
3.0352 $\pm$ 0.1953 &
3.0670 $\pm$ 0.1306 &
\underline{2.4807} $\pm$ 0.0904 &
3.0087 $\pm$ 0.1007 &
2.9035 $\pm$ 0.1042 &
\textbf{2.4719 $\pm$ 0.0351} \\

crab & 3893 & 8 &
2.4700 $\pm$ 0.1039 &
2.4073 $\pm$ 0.0741 &
2.4807 $\pm$ 0.1341 &
2.4222 $\pm$ 0.0254 &
\underline{2.3489} $\pm$ 0.1800 &
\textbf{2.2585 $\pm$ 0.0512} \\

diamond & 53940 & 9 &
5.8112 $\pm$ 0.0522 &
\underline{5.5654} $\pm$ 0.1203 &
5.9627 $\pm$ 0.0573 &
5.8208 $\pm$ 0.0609 &
5.6901 $\pm$ 0.2363 &
\textbf{5.4711 $\pm$ 0.2299} \\

forest-fires & 517 & 13 &
0.1751 $\pm$ 0.1939 &
\underline{0.1651} $\pm$ 0.0152 &
0.2121 $\pm$ 0.1313 &
0.1740 $\pm$ 0.1956 &
0.1676 $\pm$ 0.0162 &
\textbf{0.1611 $\pm$ 0.0221} \\

housing & 20640 & 9 &
4.8168 $\pm$ 0.2270 &
5.1293 $\pm$ 0.1553 &
4.8213 $\pm$ 0.1457 &
\underline{4.8148} $\pm$ 0.1256 &
4.8597 $\pm$ 0.2111 &
\textbf{4.6714 $\pm$ 0.1723} \\

insurance & 1338 & 7 &
5.2817 $\pm$ 0.2510 &
5.1103 $\pm$ 0.1670 &
\textbf{4.8053} $\pm$ 0.2061 &
4.9810 $\pm$ 0.2513 &
5.0812 $\pm$ 0.3133 &
\underline{4.8843} $\pm$ 0.1802 \\

plasma & 315 & 13 &
2.3577 $\pm$ 0.0455 &
2.2840 $\pm$ 0.1397 &
2.2409 $\pm$ 0.1554 &
\underline{2.3677} $\pm$ 0.1007 &
2.2835 $\pm$ 0.2358 &
\textbf{2.1950} $\pm$ 0.0451 \\

wine & 2554 & 11 &
0.6911 $\pm$ 0.0684 &
0.6847 $\pm$ 0.0252 &
\underline{0.6557} $\pm$ 0.0455 &
0.6911 $\pm$ 0.0822 &
0.6619 $\pm$ 0.0719 &
\textbf{0.5936} $\pm$ 0.0098 \\
\bottomrule
\end{tabular}
}
\label{tab:regression_main}
\end{table*}

Tables~\ref{tab:classification_main} and~\ref{tab:regression_main} report classification accuracy and regression RMSE respectively across all datasets and methods under a unified evaluation protocol, where all methods are paired with the same XGBoost learner and all LLM-based approaches share an identical Qwen3-8B backbone.
\paragraph{Classification.}As shown in Table~\ref{tab:classification_main}, \topo achieves the best performance on 15 of 19 datasets while maintaining consistently competitive, and often lower, variance across runs, demonstrating a favorable balance between predictive performance and search stability. Gains scale systematically with dataset complexity. On near-saturated, low-dimensional datasets (adult, tic-tac-toe), classical methods remain competitive and \topo's margins are marginal, confirming that structured multi-family search does not introduce unnecessary overhead when enumeration suffices. Gains become substantial on medium-complexity datasets requiring compositional interactions (e.g., heart, balance-scale, credit-g, eucalyptus) where \topo improves over the next-best through saturation-triggered cross-family transfer that discovers interaction features inaccessible to any single-family grammar. On large-scale datasets (diabetes), \topo achieves the best performance while LLM-FE collapses. The gap between LLM-FE and \topo is largest on datasets requiring cross-family compositions (e.g., balance-scale, heart, cmc), directly isolating heterogeneous family specialization, learned transfer topology, and adaptive prompt memory as the decisive architectural advantages.
\paragraph{Regression} As shown in Table~\ref{tab:regression_main}, \topo attains the lowest RMSE on 9 regression datasets, with improvements scaling consistently with cross-variable interaction richness: largest on interaction-driven datasets (bike: $ 28.7\%$ reduction over OCTree, airfoil: $15.7\%$ over LLM-FE) and narrowest where a single transformation family suffices (cpu, forest-fires). LLM-FE's performance gap relative to \topo widens with dataset complexity, consistent with its single-population design collapsing onto a dominant transformation family. Across both task types, \topo's lower standard deviations confirm that per-island Prompt Adaptation Memory and schema-constrained synthesis jointly reduce the instability inherent in unconstrained LLM generation.

\subsection{Generalizability}
\label{sec:generalize}
To evaluate whether \topo's effectiveness is contingent on a specific LLM backbone, we instantiate it with three models spanning a representative capability range: Qwen2.5-Coder-7B~\citep{hui2024qwen2}, a code-specialised open-source model; Qwen3-8B~\citep{yang2025qwen3}, a general-purpose open-source model; and GPT-4o-mini~\citep{hurst2024gpt}, a stronger proprietary model. All other system components remain identical across configurations.

\iffalse
To evaluate \topo's generalizability, where its effectiveness should not be contingent on a specific LLM backbone, performance should remain strong and consistent as the underlying language model varies in architecture, scale, and training methodology. This property is particularly important for real-world deployment, where access to specific models may be restricted by cost, licensing, or computational constraints. To evaluate this, we instantiate \topo with three LLM backbones spanning a representative range of capability and accessibility: (i) Code-7B, a code-specialized open-source model, (ii) Qwen3-8B, a general-purpose open-source instruction-tuned model, and (iii) GPT-4o-mini, a proprietary model with substantially stronger instruction-following and domain reasoning capabilities. All other components of \topo, implementation details and evaluation protocol remain identical across backbone configurations. XGBoost is used as the downstream predictor throughout, and results are averaged across all classification and regression datasets respectively. 
\fi
Table~\ref{tab:backbone_comparison} establishes two findings. First, \topo achieves consistently strong performance across all three backbones on both task types. The small performance variation across backbones differing in capability and training methodology confirms that \topo's core gains are driven primarily by its architectural mechanisms and slightly affected by the generative capability of LLMs. The evolutionary framework effectively compensates for weaker proposal quality through iterative fitness-guided selection. Second, classification accuracy improves monotonically with backbone capability, consistent with stronger models producing more schema-compliant programs on the first attempt and reducing wasted evaluation budget. These results confirm that \topo is backbone-agnostic in a practically meaningful sense, deployable across diverse resource constraints without architectural modification.

\begin{table}[t]
\centering
\small
\caption{Average \topo performance across all classification and regression datasets under three LLM backbones.}
\label{tab:backbone_comparison}
\begin{tabular}{lccc}
\toprule
Task & QwenCode-7B & Qwen3-8B & GPT-4o-mini \\
\midrule
Classification (Accuracy $\uparrow$) & 0.8371 & 0.8403 & 0.8468 \\
Regression (RMSE $\downarrow$)   & 2.6790 & 2.6928 & 2.6342 \\
\bottomrule
\end{tabular}
\end{table}

\subsection{Feature Transferability}
\label{sec:feat_trans}
A critical property of any feature engineering method is that discovered transformations encode dataset-intrinsic structure rather than artifacts of the search-time model. To evaluate this, we fix the feature programs evolved using XGBoost and apply the resulting augmented feature matrix without regenerating, reselecting, or modifying, to CatBoost~\citep{prokhorenkova2018catboost}, MLP~\citep{taud2017multilayer}, and TabPFN~\citep{hollmann2023tabpfn}, using Qwen2.5-Coder-7B as the backbone.

Table~\ref{tab:downstream_predictors} yields three findings. First, CatBoost achieves $0.849$ ($+1.43\%$) and RMSE $2.624$ ($-2.05\%$), confirming that \topo's features transfer without degradation across gradient-boosting variants with fundamentally different algorithmic mechanisms and are not overfit to XGBoost's split-finding algorithm. Second, MLP underperforms, attributable to the well-documented sensitivity of neural network optimizers to unnormalized skewed feature distributions~\citep{ioffe2015batch} instead of a failure of transferability. Standard normalization would be expected to substantially close this gap. Third, TabPFN achieves the best performance on both tasks, surpassing even the search-time XGBoost oracle by $4.30\%$ and $3.29\%$. This super-transfer result demonstrates \topo's programs encode richer structure than any single surrogate can fully exploit, and TabPFN's meta-learned attention mechanism is particularly well-suited to exploit the orthogonal arithmetic, temporal, rank-based, and aggregation representations that \topo's family-specialized islands produce. These results confirm that transferability is a stable, task-agnostic property of \topo's feature programs, with improvement magnitude scaling with each predictor's capacity to exploit structured low-redundancy representations.

\begin{table}[t]
\centering
\setlength{\tabcolsep}{3pt}
\caption{Average downstream performance using the same engineered feature set generated by \topo with XGBoost as the default predictor.}
\label{tab:downstream_predictors}

\begin{tabular}{lcccc}
\toprule
Task & XGB & CatBoost & MLP & TabPFN \\
\midrule
Classification (Accuracy $\uparrow$)
& 0.837
& 0.849
& 0.816
& \textbf{0.873} \\

Regression (RMSE $\downarrow$)
& 2.679
& 2.624
& 2.993
& \textbf{2.591} \\
\bottomrule
\end{tabular}

\end{table}

\subsection{Feature Diversity}
\label{sec:feature_disversity}
To evaluate the structural quality and informational richness of the feature sets produced by each method, we define two geometrically complementary diversity metrics. The \textit{Mean Pairwise Output Correlation} (MPOC) measures local pairwise redundancy as the average absolute Spearman rank correlation between all pairs of output vectors in the accepted engineered feature set $\Phi$:
\begin{equation}
\mathrm{MPOC}(\Phi)
=
\frac{2}{|\Phi|(|\Phi|-1)}
\sum_{i<j}
\left|
\rho_{\mathrm{sp}}
\big(
\phi_i(\mathbf{X}),
\phi_j(\mathbf{X})
\big)
\right|,
\label{eq:mpoc}
\end{equation}
where $\rho_{\mathrm{sp}}$ denotes the Spearman rank correlation evaluated on the training set. Lower MPOC indicates that the accepted features carry less redundant information, with output vectors spanning a larger subspace of $\mathbb{R}^{n}$ instead of collapsing into a low-rank correlated block.

\textit{Effective Rank} (EffRank) measures global dimensional coverage from the normalized singular-value spectrum of the engineered feature matrix:
\begin{equation}
\mathrm{EffRank}(\Phi)
=
\exp\left(
-\sum_{k=1}^{r} \tilde{\sigma}_k \log \tilde{\sigma}_k
\right),
\qquad
\tilde{\sigma}_k
=
\frac{\sigma_k}{\sum_{j=1}^{r} \sigma_j}
\label{eq:effrank}
\end{equation}
where $\{\sigma_k\}_{k=1}^{r}$ are the singular values. Higher EffRank indicates broader coverage of independent feature directions. We report EffRank normalized by $|\Phi|$ for comparability across methods with different numbers of accepted features.

\begin{table}[t]
\centering
\small
\caption{Average MPOC and Effective Rank over all classification and regression datasets across all methods.}
\label{tab:mpoc_effirank}
\begin{tabular}{lcccc}
\toprule
\multirow{2}{*}{Method} 
& \multicolumn{2}{c}{Classification} 
& \multicolumn{2}{c}{Regression} \\
\cmidrule(lr){2-3} \cmidrule(lr){4-5}
& MPOC $\downarrow$ & EffRank$\uparrow$  & MPOC$\downarrow$  & EffRank$\uparrow$ \\
\midrule
OpenFE   & 0.421 & 0.480 & 0.378 & 0.448 \\
AutoFeat & 0.513 & 0.330 & 0.486 & 0.309 \\
FeatLLM  & 0.581 & 0.394 & 0.462 & 0.389 \\
CAAFE    & 0.479 & 0.567 & 0.382 & 0.412 \\
LLMFE    & 0.309 & 0.699 & 0.320 & 0.631 \\
\textbf{\topo}   & \textbf{0.261} & \textbf{0.832} & \textbf{0.232} & \textbf{0.680}\\
\bottomrule
\end{tabular}

\end{table}

Table~\ref{tab:mpoc_effirank} reveals that \topo achieves the lowest MPOC and highest EffRank across both classification and regression, establishing joint optimality in redundancy suppression and subspace coverage that no competing method achieves. The per-method profiles expose mechanistically distinct failure modes: AutoFeat attains the highest MPOC and lowest EffRank because polynomial enumeration generates strongly monotone-related features that collapse the singular value spectrum into few dominant directions. FeatLLM similarly suffers high MPOC and low EffRank through thematic clustering of stateless LLM proposals that are semantically coherent but informatively redundant. CAAFE and OpenFE occupy an intermediate regime where iterative feedback and feature boosting partially suppress redundancy but remain confined to a single operator grammar, precluding high EffRank. LLM-FE improves substantially yet is ultimately constrained by its homogeneous, grammar-agnostic islands that concentrate the feature matrix within a single undifferentiated operator subspace. \topo's joint advantage is explained by two complementary mechanisms absent from all prior work: an explicit within-search redundancy filter that prevents correlated clusters from accumulating (driving MPOC down), and heterogeneous family schema combined with cross-family hybrid synthesis that import structurally orthogonal templates from precursor islands into the target grammar (driving EffRank up), together constituting the geometric signature of \topo's architectural superiority.

\subsection{Topology Graph Specialization}
\label{sec:learning_effec}
To evaluate whether \topo's topology graph acquires task-specific knowledge about cross-family transfer utility over the course of search, we define the \textit{Topology Graph Specialization Score} (TGSS). Let $W^{(t)} \in \mathbb{R}^{M \times M}$ denote the transfer-weight matrix of the topology graph at generation $t$, where each off-diagonal entry $w_{j \to i}^{(t)} \in [-1,1]$ encodes the learned utility of island $j$ as a precursor for saturated island $i$ and is updated according to Eq.~\ref{eq:weightupdate}. TGSS is defined as the empirical variance of the off-diagonal transfer weights:
\begin{equation}
\mathrm{TGSS}(t)
=
\frac{1}{M(M-1)}
\sum_{\substack{j,i=1 \\ j \neq i}}^{M}
\left(
w_{j \to i}^{(t)} - \bar{w}^{(t)}
\right)^2
\label{eq:tgss}
\end{equation}
where the mean off-diagonal weight is defined as
\begin{equation}
\bar{w}^{(t)}
=
\frac{1}{M(M-1)}
\sum_{\substack{j,i=1 \\ j \neq i}}^{M}
w_{j \to i}^{(t)}.
\label{eq:mean_transfer_weight}
\end{equation}

TGSS quantifies the structural divergence of the weight matrix from its uniform initialization: $TGSS(0)=0$ exactly when all weights are equal, and increases as the graph develops differentiated preferences over island pairs. A high TGSS indicates that the topology graph has identified strongly productive and strongly unproductive transfer directions specific to the current task, concentrating transfer routing accordingly. A near-zero TGSS indicates that all island pairs yield approximately equal transfer utility and the graph remains close to its uninformative prior. TGSS directly measures whether the adaptive topology mechanism is acquiring task-specific knowledge that distinguishes \topo from fixed or random transfer alternatives.
\begin{figure}[h]
    \centering
    \includegraphics[width=0.49\textwidth]{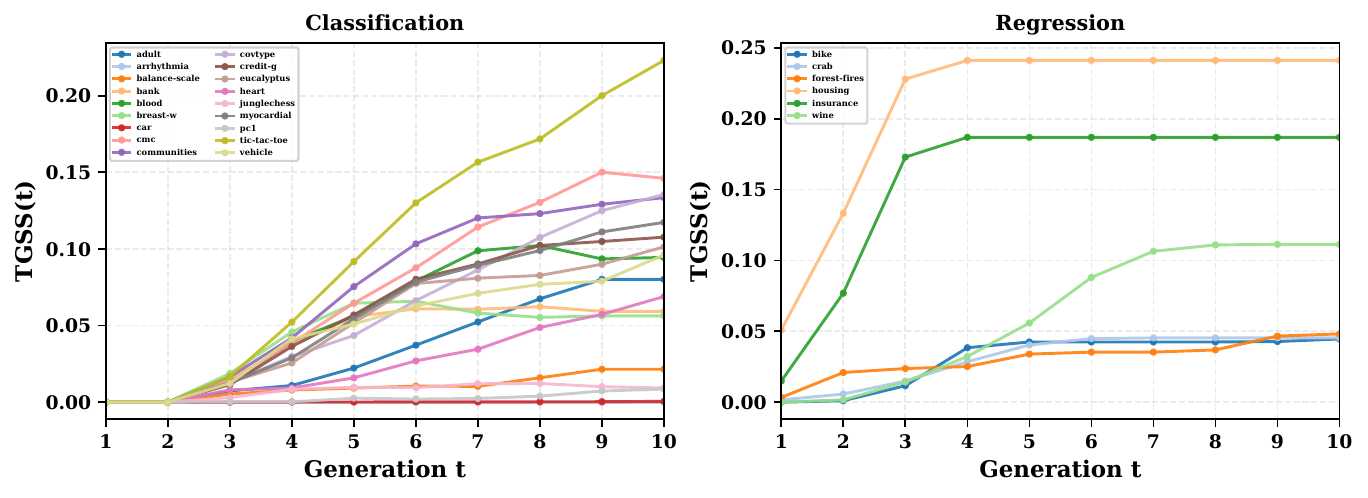}
    \caption{TGSS trajectories across all classification (left) and regression (right) datasets over 10 generations. }
    \label{fig:tgss}
\end{figure}

Figure~\ref{fig:tgss} reports TGSS trajectories across all classification and regression datasets over 10 generations. All datasets begin at $TGSS(0)=0$ and increase monotonically, confirming that the topology graph consistently specializes away from its uniform prior as transfer events accumulate --- a necessary precondition for adaptive routing to confer any advantage over random precursor selection. Three patterns emerge from the joint analysis. First, TGSS exhibits substantial inter-dataset variance at convergence, spanning $[0.00,0.22]$ for classification and $[0.04,0.24]$ for regression, confirming that the degree of learnable cross-family utility structure is genuinely dataset-dependent instead of a fixed algorithmic property. This validates \topo's adaptive design: a fixed topology would be suboptimal for all datasets, while the learned graph adapts to the specific inter-family utility structure of each task. Second, datasets with near-zero TGSS throughout, notably tic-tac-toe and blood dataset (classification) are precisely those where \topo's performance advantage over single-family baselines is smallest, establishing TGSS as a practically useful predictive indicator of when cross-family transfer is likely to be beneficial. Conversely, vehicle (classification, $TGSS\approx 0.22$) and forest-fires (regression, $TGSS\approx 0.24$) achieve the highest specialization, consistent with their heterogeneous feature interactions across multiple operator families that produce strongly differentiated island pair utilities. Third, regression datasets exhibit faster early plateau behavior. Most curves flatten after generation 3--4. while classification curves continue growing through generation 10. This asymmetry reflects that regression fitness signals are smoother and more stable, enabling the bandit-style update rule to reach confident weight estimates in fewer transfer trials, while classification tasks present noisier utility landscapes that require sustained transfer sampling to resolve inter-family utility differences. This observation motivates task-type-specific transfer budget scheduling as a direction for future work. 

\subsection{Ablation Study}
\label{sec:ablation}
To quantify the individual contribution of each architectural component in \topo, we conduct a systematic ablation study by progressively removing or modifying key mechanisms while holding all other components fixed. We evaluate ten variants organised into four groups corresponding to the main architectural decisions: island decomposition, migration strategy, transfer mechanism, and saturation detection.

\textbf{Island decomposition:} (i) \textit{Single Pool} removes island decomposition and migration, reducing \topo to single-population search; (ii) \textit{Random Island} partitions the population into islands without family-aware assignment. \textbf{Migration strategy:} (iii) \textit{w/o Inter-island Exchange} retains family decomposition but disables cross-island migration; (iv) \textit{Static Migration} replaces saturation-triggered transfer with fixed-interval migration; (v) \textit{Stochastic Migration} selects precursor islands uniformly at random. \textbf{Transfer mechanism:} (vi) \textit{Direct Transfer} copies programs from the precursor island without hybrid synthesis; (vii) \textit{Unguided Hybridisation} enables hybrid synthesis with randomly selected precursors, disabling topology guidance; (viii) \textit{w/o Hybridisation} preserves topology-guided precursor selection but removes LLM-mediated synthesis. \textbf{Saturation detection:} (ix) \textit{Correlation-Only} replaces the composite saturation criterion with a single output-correlation signal; (x) \textit{Multi-signal Saturation} restores full multi-signal adaptive saturation detection without other full-system components.

\begin{figure}[h]
    \centering
    \includegraphics[width=0.495\textwidth]{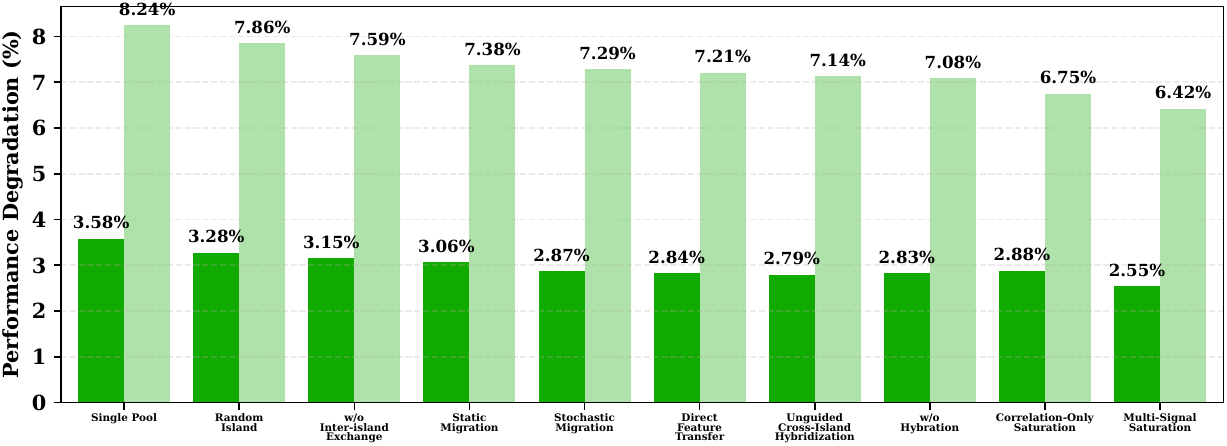}
    \caption{Ablation study of \topo. Each bar reports the performance degradation relative to the full \topo. Results are averaged across all datasets.}
    \label{fig:ablation}
\end{figure}

Figure~\ref{fig:ablation} summarizes performance degradation across all variants on both task types. The largest drop occurs for \textit{Single Pool}, confirming that unstructured single-population search is the primary bottleneck: without island decomposition, search collapses toward dominant transformation patterns, causing premature convergence and reduced diversity. The improvement from \textit{Random Island} to \textit{w/o Inter-island Exchange} further demonstrates that family-aware decomposition contributes meaningful inductive structure beyond simple partitioning, improving local optimization efficiency by conditioning LLM proposals on a homogeneous operator context. Migration ablations show that transfer effectiveness depends critically on both timing and precursor quality. \textit{Static Migration} wastes transfer budget when local search remains productive, while \textit{Stochastic Migration} degrades further by introducing structurally unrelated transformations that disrupt locally adapted populations. Transfer mechanism ablations confirm the necessity of compositional synthesis: \textit{Direct Transfer} yields limited gains as copied programs remain constrained by their source family's inductive bias. LLM-mediated synthesis overcomes this by generating programs spanning multiple families that are unreachable by intra-island search alone. \textit{Unguided Hybridization} remains suboptimal despite enabling synthesis, demonstrating that topology guidance and hybrid synthesis are synergistic --- learned precursor selection ensures hybridization targets complementary family pairs, maximizing the probability of introducing non-redundant predictive structure. Saturation detection ablations confirm that multi-signal stagnation detection is essential. \textit{Correlation-Only Saturation} underperforms because output correlation alone cannot distinguish genuine stagnation from transient redundancy --- an island may stagnate in performance while maintaining low correlation, or exhibit high correlation from isolated redundant proposals while remaining broadly productive. The multi-signal composite criterion correctly discriminates these cases, triggering transfer only when local search is genuinely exhausted across all dimensions of diversity and performance.

\section{Related Work}
\label{sec:related}

\subsection{Automatic Feature Engineering}
\label{sec:afe}
Automated feature engineering (AutoFE) has evolved through three broad paradigms. \emph{Search-based methods} treat feature construction as a combinatorial optimization problem, through genetic programming~\citep{olson2016tpot}, exhaustive~\citep{horn2019autofeat} and greedy search~\citep{zhang2023openfe}. These methods are computationally expensive and domain-agnostic without semantic guidance. \emph{LLM-guided methods} use pretrained LMs to propose features~\citep{han2024large, ko2025ferg, nam2024optimized} from task descriptions and dataset metadata~\citep{hollmann2023large}. Recent extensions apply Tree-of-Thought reasoning~\citep{zhang2025dynamic}, multi-agent frameworks~\citep{ouyang2025fela}, and ReAct-based agents~\citep{burghardt2026famose} to guide feature discovery. However, these methods reply on prompting, leading to repeated proposals, no dataset-specific adaptation, and no cross-family composition. \emph{LLM-evolutionary methods} combine LLM-based generation with evolutionary search~\citep{abhyankar2025llm, gong2025evolutionary, batista2025embedding}. While these methods improve over static prompting, they operate over a single population without structural decomposition of the program space, precluding cross-family hybridization and leaving the search vulnerable to premature convergence within a dominant transformation family. \topo addresses these limitations directly through multi-island decomposition, prompt adaptation memory, and topology-guided cross-family transfer.

\subsection{LLMs for Tabular Data Learning}
\label{sec:llm4tabular}
Recent advances in tabular learning have converged toward two main paradigms. \emph{Tabular foundation models} pursue cross-table transferability through large-scale pretraining, ranging from Bayesian inference approximation via Prior-Data Fitted Networks~\citep{hollmann2023tabpfn, hollmann2025tabpfn} and architecture-level scaling for in-context learning~\citep{gardner2024large, arazi2026tabstar, qu2025tabicl}, to retrieval-augmented inference without task-specific fine-tuning~\citep{ma2025tabdpt}. \emph{LLM-based tabular methods} repurpose pretrained LMs for few-shot prediction~\citep{hegselmann2023tabllm, dinh2022lift}, semantic generalization via textual metadata~\citep{ye2024towards, kim2024carte, yanmaking}, and auxiliary tasks including data augmentation~\citep{zhang2023generative, borisovlanguage}, data cleaning~\citep{bendinelli2025exploring}, and feature generation~\citep{hollmann2023large}. Despite these advances, existing approaches largely rely on per-task adaptation and lack principled search mechanisms over the feature space, which \topo is specifically designed to provide.

\subsection{LLM-guided Evolutionary Search}
A growing body of work integrates LLMs as evolutionary operators across prompt optimization~\citep{guo2023evoprompt, fernandopromptbreeder, suzgun2026dynamic, agrawalgepa}, program and algorithm discovery~\citep{vsurinaalgorithm, assumpccao2025codeevolve, novikov2025alphaevolve, sharma5openevolve, lange2025shinkaevolve}, and scientific discovery~\citep{shojaeellm, yang2025heuragenix, chen2026molevolve, abhyankar2026llema}. These systems vary primarily in population dynamics, employing MAP-Elites~\citep{novikov2025alphaevolve}, island-based evolution~\citep{assumpccao2025codeevolve}, Pareto frontier optimization~\citep{agrawalgepa}, and diversity-driven selection~\citep{sharma5openevolve}, with more advanced frameworks incorporating co-evolution and meta-reflection~\citep{liuevolution, ye2024reevo} or iterative agent evolution~\citep{yuan2025evoagent, shangagentsquare}. Despite their effectiveness, existing methods treat the search space as structurally uniform and do not incorporate semantic decomposition, task-adaptive proposal conditioning, or saturation-triggered transfer, all of which are central to \topo.

\section{Conclusion}
\label{sec:conclusion}
We present \topo, a LLM-guided framework for automated feature engineering that models search as graph-structured multi-island evolution over heterogeneous transformation families, addressing three fundamental limitations of existing methods: homogeneous search dynamics, stateless LLM querying, and rigid migration topology. Family-specialised islands preserve transformation diversity by construction; a learned directed topology graph routes cross-family transfer toward empirically productive island pairs; and saturation-triggered LLM-mediated hybrid synthesis realises transfer as compositional program generation precisely when local search is exhausted. Across 19 classification and 10 regression benchmarks, \topo consistently outperforms all baselines, with the largest gains on datasets requiring cross-family feature interactions. Three purpose-built metrics confirm the mechanistic basis: MPOC and EffRank demonstrate lower-redundancy, higher-coverage feature sets than all competing methods, TGSS confirms that the topology graph reliably acquires task-specific transfer knowledge, with specialisation degree correlating directly with performance gains over single-family baselines. Feature programs transfer reliably across architecturally distinct predictors, and remain stable across LLM backbones of varying capability, confirming that gains are driven by architectural mechanisms rather than any particular model. These results collectively establish that inter-family complementarity is a learnable and exploitable quantity that existing AutoFE methods leave entirely untapped, opening promising directions for automated family schema induction, task-type-specific transfer scheduling, and extension to multi-modal tabular settings.

\newpage
\bibliographystyle{unsrtnat}
\bibliography{references}

\newpage
\textbf{Appendix}
\appendix

\section{Comparison with LLM-based Baseline}
\label{app:comparison}
We compare and summarize the differences between \topo and other LLM-based methods.

\textbf{(i) Exploration Strategy.}
CAAFE explores greedily and sequentially: it proposes one candidate feature at a time, accepts or rejects it based on validation performance, and repeats for a fixed number of iterations. FeatLLM runs many independent LLM queries in parallel, but the trials never interact with one another. OCTree follows a single-trajectory iterative optimization, refining one rule at a time before moving on to the next feature. LLM-FE performs evolutionary search over a single population that is split into island buffers, with mutation and crossover realized implicitly through prompt-conditioned generation. \topo evolves multiple family-specialized islands in parallel, each with explicit LLM-driven mutation and crossover operators, and adds saturation-triggered cross-island exploration on top. The essential delta is that every baseline explores either one path (CAAFE, OCTree), independent paths (FeatLLM), or interchangeable paths (LLM-FE), whereas \topo is the only method whose paths are differentiated by inductive bias and actively coordinated, each island knows what kind of feature it is searching for and learns which other island to borrow from.

\textbf{(ii) Search Space.}
CAAFE's search space consists of unconstrained code expressions, which in practice collapse into simple arithmetic combinations. FeatLLM deliberately restricts its space to per-class threshold and membership rules over individual features. OCTree searches over natural-language rules that are subsequently converted into code, with no pre-defined operator space. LLM-FE searches the space of full feature-transformation programs, but over one undifferentiated grammar. \topo also searches full transformation programs, but partitions the space into semantically typed subspaces, and additionally makes the joint compositional spaces between families reachable through hybrid synthesis. \topo is the only method that decomposes the program space into typed subspaces and treats cross-family compositions as explicit, deliberately-triggered targets: LLM-FE has the same raw expressiveness but one flat grammar, FeatLLM deliberately limits itself to simple rules, and CAAFE is unconstrained in principle yet simple in practice.

\textbf{(iii) Memory Management.}
CAAFE remembers only the accepted features and the most recent execution error, so rejected features leave no trace and near-duplicates of failed ideas can recur. FeatLLM is fully stateless, each trial is an independent LLM call with no memory across trials. OCTree retains the full optimization trajectory of prior rules, their scores, and the associated reasoning, re-sent to the LLM at every call. LLM-FE maintains a buffer of high-scoring programs, but it stores winners only. \topo maintains a dual memory: an elite archive of exemplar programs, plus a Prompt Adaptation Memory that summarizes both the accepted and the rejected history of each island into a compact natural-language ``prefer/avoid" signal, continuously updated during search. 

\textbf{(iv) Elite Management.}
CAAFE has no explicit archive: the currently accepted feature set effectively is the elite. FeatLLM keeps no elites at all, the final ensemble simply averages the outputs of all trials. OCTree keeps only the best-scoring rule per feature, with no population behind it. LLM-FE maintains a score-clustered elite buffer, sampled to provide in-context
demonstrations, and deduplicates programs by their performance signature. \topo maintains a bounded per-island elite archive that enforces structural novelty, candidates too syntactically similar to existing members are discarded, and this archive doubles as both the in-context demonstration pool and the donor material for cross-island hybridization.
\topo's elites are deduplicated by structure, keeping the archive behaviorally and syntactically diverse, which matters because elites serve as crossover and hybridization parents, so redundant parents would produce redundant offspring.

\textbf{(v)  Optimization ObjectivM.}
CAAFE feeds back a scalar performance change plus any execution errors. FeatLLM provides no iterative feedback, the LLM sees only few-shot examples. OCTree feeds back validation scores together with decision-tree reasoning expressed in natural language, giving the LLM structural knowledge about the dataset. LLM-FE feeds back validation scores alongside high-scoring programs as in-context exemplars. \topo feeds back the fitness score, elite exemplars, and the distilled preference signal from its prompt memory. The key distinction within the ``rich feedback" camp is that OCTree's decision tree summarizes the data, whereas \topo's prompt memory summarizes the search experience, the two are orthogonal forms of structured feedback, and \topo steers the LLM's proposal distribution with an explicit, cumulative account of which regions of the search space have proven productive or dead.

\textbf{(vi)  Feedback Signal to the LLM.}
CAAFE maximizes validation classification performance. FeatLLM minimizes the loss of a simple model fitted on its binary rule features. OCTree minimizes validation loss for each rule independently. LLM-FE maximizes the validation score of the best program found. \topo alone optimizes an explicitly multi-term objective that combines downstream performance, a redundancy penalty over pairwise feature correlations, a stability term rewarding consistency across validation folds, and a computational cost term. Every baseline treats predictive performance as the sole criterion, whereas \topo formalizes feature quality as a multi-criteria property, and this is precisely why it uniquely wins on both redundancy and subspace-coverage metrics instead of on accuracy alone.

\textbf{(vii)  Parallelism and Efficiency.}
CAAFE and OCTree are inherently sequential, each iteration depends on the previous one's outcome. FeatLLM's independent trials are trivially parallel. LLM-FE's islands are parallelizable in principle. \topo is parallel by design: island evaluations are mutually independent, so wall-clock time per generation is governed by the slowest island rather than the total number of candidates, while saturation-triggered (rather than continuous) transfer keeps coordination overhead negligible.

\section{Prompt Design}
\label{app:prompts}

This section documents the prompts of \topo, organized by function. 

\subsection{Feature-Understanding Call}
\label{app:understanding}

\begin{verbatim}
###
<Role>
You are an scientist working on feature engineering program generation.

###
<Task>
First understand the dataset features and task.
Explicitly reason about relationships between features and the prediction target.
Assume all input features are usable; do not exclude or de-prioritize any column.
Target: {task_description}
Columns: {comma-separated columns}

###
<Output>
Return concise bullets:
1) likely feature-to-target relationships (direction, nonlinearity, thresholds, monotonic effects
when plausible)
2) likely useful feature-to-feature interactions/ratios/aggregations that could improve
target prediction
3) concrete transformation ideas for each major feature group (numeric, categorical, mixed)
\end{verbatim}

\subsection{Generation Context}
\label{app:gen_cont}
This prompt is shared generation context that prepended to initialize, mutation, crossover, and hyrbid. 

\begin{verbatim}
###
<Objective>
Primary mission: "extract, transform, and select variables from raw data by generating feature
engineering Python codes based on the analysis, making machine learning models more accurate and 
efficient."
Generation objective: create engineered features that capture both:
- feature-to-target relationships
- feature-to-feature relationships relevant to target prediction
Avoid arbitrary transformations that are not tied to target-relevant signal.
First include a concise one-sentence docstring describing the transformation intent.

\end{verbatim}

\subsection{Generation Rules}
\label{app:gen_roles}
\begin{verbatim}
###
<Role>
You are a data scientist with expert knowledge about the provided dataset.
Your role is to identify and engineer informative features to solve the <Task> effectively.

###
<Instructions>
- Return Python code lines only.
- Ensure the code is executable. 
- Write the BODY of a function `generated_feature(df)`.
- Do NOT use placeholders/undefined names like member1, feature1, x, y.
- Do NOT create toy/sample dataframes; operate on provided df only.
- Avoid providing any additional descriptions outside code.
- Use valid operator calls with dataframe series arguments.
Allowed operators for this family: {family_guide}
Current family: {family_name}

\end{verbatim}

\subsection{Initialization Prompt}
\label{app:init_prompt}

The initialization prompt seeds each island with a strong baseline program:
\begin{verbatim}
{prompt_context}{generation_rules}
Produce ONE initial '{family_name}' feature program.
Use at least two source columns, or an axis=1 aggregation
across many columns.
Ground it in the analysis notes; make it a strong,
self-contained baseline for this family.
\end{verbatim}

\subsection{Mutation Prompt}
\label{app:mutation_prompt}

The mutation prompt modifies a single parent program conditioned on PAM signals:
\begin{verbatim}
{prompt_context}{generation_rules}
Family: {family_name}
Parent program:
{parent.code}

Preferred patterns (recently rewarded): {prefer}
Avoid patterns (recently rejected/redundant): {avoid}

Mutate the parent into a distinct, stronger '{family_name}'
feature. Change its structure meaningfully swap an operator, add a stabilizing transform, or 
bring in another column so the output is not merely a rescaled version of the parent.
\end{verbatim}

The \texttt{\{prefer\}} and \texttt{\{avoid\}} fields are populated from the prompt-memory vector $\rho_i^{(t)}$, providing explicit directional guidance derived from the island's accumulated accept/reject history (Eq.~\ref{eq:pamupdate}).

\subsection{Crossover Prompt}
\label{app:crossover_prompt}

The crossover prompt synthesizes a child from two parents within the same island:
\begin{verbatim}
{prompt_context}{generation_rules}
Family: {family_name}
Parent A:
{p1.code}

Parent B:
{p2.code}

Combine the most useful structural idea from each parent into
one new '{family_name}' feature.
Do not concatenate them verbatim; synthesize a coherent signal
that improves on both and stays low-redundancy.
\end{verbatim}

\subsection{Cross-Island Hybrid Synthesis Prompt}
\label{app:hybrid_prompt}

The cross-island hybrid prompt is triggered by saturation detection and generates programs in the joint composition space $\mathcal{P}_i \circ \mathcal{P}_j$:
\begin{verbatim}
{prompt_context}{generation_rules}
TARGET FAMILY: {target.family_name}
DONOR FAMILY: {donor.family_name}

Target parent (stay in this family's style):
{p_t.code}

Donor parent (borrow ONE idea from a different family):
{p_d.code}

The target island has stagnated. Import a single useful
structural idea from the donor - an operator, a
normalization, or a grouping strategy, and re-express it
as a '{target.family_name}' feature.
Do not copy either parent. The result must be robust,
non-trivial, and structurally novel relative to the target
island's existing features.
\end{verbatim}

\end{document}